\documentclass[conference]{IEEEtran}
\IEEEoverridecommandlockouts
\usepackage{cite}
\usepackage{amsmath,amssymb,amsfonts}
\usepackage{algorithmic}
\usepackage{graphicx}
\usepackage{textcomp}
\usepackage{xcolor}
\def\BibTeX{{\rm B\kern-.05em{\sc i\kern-.025em b}\kern-.08em
    T\kern-.1667em\lower.7ex\hbox{E}\kern-.125emX}}

\usepackage{xspace}
\usepackage{tabularx}
\usepackage{booktabs} 
\usepackage{pifont} 
\usepackage{colortbl} 
\usepackage[ruled,vlined,linesnumbered,commentsnumbered]{algorithm2e}
\usepackage{multirow}
\usepackage{CJKutf8} 
\usepackage{float}
\usepackage{url}

\usepackage{amsthm}

\newtheoremstyle{definition_bold} 
  {3pt} 
  {3pt} 
  {\normalfont} 
  {\parindent} 
  {\itshape} 
  {} 
  {.5em} 
  {\thmname{#1}\thmnumber{ #2}:\thmnote{ {\bfseries\upshape (#3)}}} 
\theoremstyle{definition_bold}

\newtheorem{definition}{Definition}

\newcommand{\unitdb}{\texttt{DimUnitKB}\xspace}
\newcommand{\unitsys}{\texttt{DimKS}\xspace}
\newcommand{\uniteval}{\texttt{DimEval}\xspace}
\newcommand{\dimperc}{\texttt{DimPerc}\xspace}

\SetCommentSty{mycommfont}

\begin{document}

\title{Enhancing Quantitative Reasoning Skills of Large Language Models through Dimension Perception

\noindent \thanks{\IEEEauthorrefmark{1}Corresponding author. Yanghua Xiao is also a member of Research Group of Computational and AI Communication at Institute for Global Communications and Integrated Media, Fudan University.}
}




\author{
\IEEEauthorblockN{
Yuncheng Huang\IEEEauthorrefmark{2},
Qianyu He\IEEEauthorrefmark{2},
Jiaqing Liang\IEEEauthorrefmark{3}\IEEEauthorrefmark{1},
Sihang Jiang\IEEEauthorrefmark{2},
Yanghua Xiao\IEEEauthorrefmark{2}\IEEEauthorrefmark{1},
Yunwen Chen\IEEEauthorrefmark{4}}
\IEEEauthorblockA{\IEEEauthorrefmark{2}Shanghai Key Laboratory of Data Science, School of Computer Science, Fudan University}
\IEEEauthorblockA{\IEEEauthorrefmark{3}School of Data Science, Fudan University \IEEEauthorrefmark{4}DataGrand Co., LTD.}
\IEEEauthorblockA{\{yunchenghuang22, qyhe21\}@m.fudan.edu.cn, \{liangjiaqing, shawyh\}@fudan.edu.cn}
\IEEEauthorblockA{tedsihangjiang@gmail.com, chenyunwen@datagrand.com}
}




\maketitle

\begin{abstract}
Quantities are distinct and critical components of texts that characterize the magnitude properties of entities, providing a precise perspective for the understanding of natural language, especially for reasoning tasks.
In recent years, there has been a flurry of research on reasoning tasks based on large language models (LLMs), most of which solely focus on numerical values, neglecting the dimensional concept of quantities with units despite its importance.
We argue that the concept of dimension is essential for precisely understanding quantities and of great significance for LLMs to perform quantitative reasoning.
However, the lack of dimension knowledge and quantity-related benchmarks has resulted in low performance of LLMs.
Hence, we present a framework to enhance the quantitative reasoning ability of language models based on dimension perception.
We first construct a dimensional unit knowledge base (\unitdb) to address the knowledge gap in this area.
We propose a benchmark \uniteval consisting of seven tasks of three categories to probe and enhance the dimension perception skills of LLMs.
To evaluate the effectiveness of our methods, we propose a quantitative reasoning task and conduct experiments.
The experimental results show that our dimension perception method dramatically improves accuracy (43.55\%$\rightarrow$50.67\%) on quantitative reasoning tasks compared to GPT-4.
\end{abstract}


\section{Introduction}
Quantities, an essential part of text, are utilized to accurately characterize the magnitude attributes of entities, wherein they abound in the corpus with distinctive and vital significance~\cite{thawani-etal-2021-representing}.
In many scenarios, the concepts of numerical value and quantity are often confused. 
As delineated in~\cite{thawani-etal-2021-representing}, \textbf{abstract values} are values that consist solely of numerical parts, while \textbf{grounded values} encompass both numerical and unit parts, with the latter referred to as \textbf{quantities} in this paper.
It is essential for language models to grasp the nature of quantities in order to precisely understand natural language texts, especially for reasoning tasks~\cite{ran2019numnet,lin-etal-2020-birds,Mishra2020TowardsQF}.
For instance, through precise unit conversion and value comparison, language models should be able to deduce the conclusion that \textit{``LeBron James is taller than Stephen Curry''} when the text provides information such as \textit{``LeBron James's height is 2.06 meters and Stephen Curry's height is 188 cm.''}

In recent years, large language models (LLMs)~\cite{brown2020language,chowdhery2022palm,chung2022scaling,sanh2021multitask} such as ChatGPT have made significant advancements in a variety of reasoning tasks, such as arithmetic reasoning, commonsense reasoning,  symbolic reasoning, etc~\cite{huang2022towards,wei2022chain}.
Extensive research has shown that LLMs are capable of extracting implicit knowledge from pretrained corpus and help enhance the performance on reasoning tasks~\cite{li-etal-2022-systematic}.
However, language models exhibit limited capabilities when it comes to mathematical reasoning tasks, particularly those involving quantities~\cite{park-etal-2022-language}.
We argue that quantity-related reasoning tasks are of significant importance due to their high domain-specific applicability in fields such as mathematics, finance, engineering, physics, biomedical sciences, and chemistry.
In this paper, we refer to reasoning tasks that necessitate thorough analysis of quantities as \textbf{quantitative reasoning} tasks.

\begin{figure}
    \centering
    \includegraphics[width=0.46\textwidth, height=50mm]{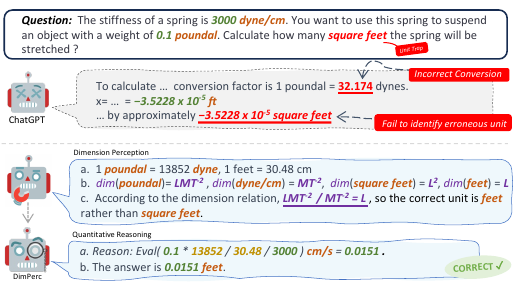}
    \caption{An example of quantitative reasoning through dimension perception. ChatGPT (October 31, 2023) failed to identify the incorrect units in the question due to a lack of understanding of dimensional concepts, leading to erroneous inferences. Our approach derives accurate quantities through dimensional knowledge. Dimension is strictly defined in Section~\ref{sec:preliminary}.}
    \label{fig:intro}
\end{figure}

\begin{figure*}[t]
    \centering
    \includegraphics[width=1\textwidth,height=60mm]{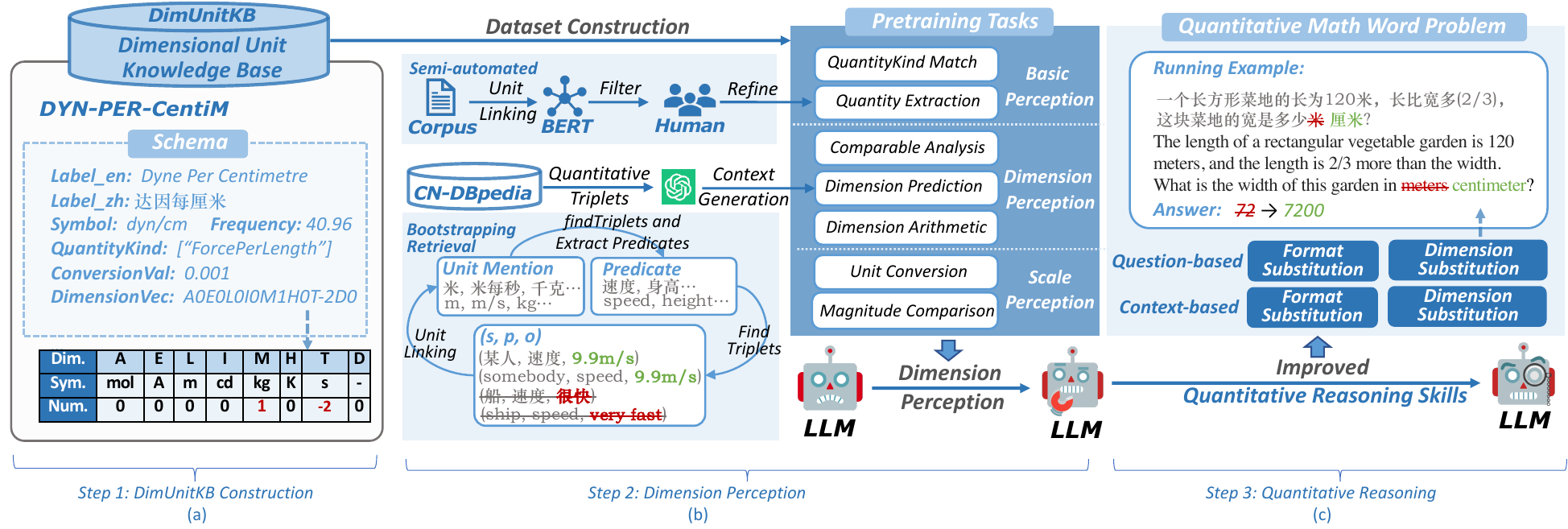}
    \caption{The framework for enhancing quantitative reasoning skills through dimension perception. (a) Step 1 involves the construction of a dimensional unit knowledge base (\unitdb). (b) Step 2 proposes and develops seven pretraining tasks of three categories for dimension perception, with corresponding datasets assembled through semi-automated and bootstrapping retrieval methods. (c) Step 3 improves performance by utilizing quantity-oriented data augmentation for tasks under quantitative reasoning.}
    \label{fig:overview}
\end{figure*}

Since language models are based on probabilistic principles, it is challenging to carry out rigorous reasoning tasks like quantitative reasoning~\cite{lee2023teaching}.
More importantly, we argue that one of the major challenges in quantitative reasoning lies in the perception of \textbf{dimension}.
This is because dimensions are the fundamental attributes of quantities, which reflects the relationships among different quantities~\cite{fourier1822theorie}.
We formally define the concept of dimension in Section~\ref{sec:preliminary}.
As shown in Fig.~\ref{fig:intro}, the dimension of ``$poundal$'' and ``$dyn/cm$'' are ``$LMT^{-2}$'' and ``$MT^{-2}$'' respectively.
In this example, we set up a unit trap in the question. ChatGPT used incorrect unit conversions in this question and is unable to recognize the unit trap.
By applying our methods, such errors can be identified, leading to correct reasoning outcomes.

\subsection{Limits of Previous Solutions}
Although there are numerous studies involving numerical reasoning~\cite{xie2019goal, dua-etal-2019-drop, ggb2020injecting}, most of them center on abstract numerical values, neglecting quantities that encompass a wide range of units, despite their importance.
Models derived from these studies are insufficient for quantity comprehension, as they disregard the distinction between units and simply reason by abstract value, leading to erroneous results. 

There are only a few studies that involve mathematical reasoning concerning quantities\cite{thawani-etal-2021-numeracy,lin2020birds,park-etal-2022-language}.
UoM~\cite{park-etal-2022-language} investigates the language model's ability to comprehend quantities and further substantiates the inadequacy of language models in understanding quantities, especially units.
However, these studies fall short in their coverage of existing frequently used units.
Furthermore, due to a scarcity of annotated datasets that cover a wide range of quantities, existing language models still struggle with comprehensively understanding the quantities. 
Crucially, these models do not employ the essential concept of dimension, leading to deficiencies in quantitative reasoning scenarios.

\subsection{Our ideas and Contributions}

In this paper, we focus on \textbf{quantities} in natural language texts.
Our goal is to \textit{enhance the quantitative reasoning skills of LLMs through dimension perception}. 
To achieve the goal, we need the following efforts.

First, we need to address the lack of knowledge of the language models for the various units.
Previous probing research suggests that it is difficult for language models to acquire the knowledge of unit conversion from corpora~\cite{lin-etal-2020-birds}.
Furthermore, existing knowledge bases or graphs mainly cover facts, concepts, or entities ~\cite{liu2004conceptnet,auer2007dbpedia,bollacker2008freebase,vrandevcic2014wikidata}, which are insufficient for understanding quantities, particularly the concept of dimension.
To tackle these issues, we first construct a dimensional unit knowledge base (\unitdb), which stores basic information, including dimension knowledge of units.
We develop a unit linking module on top of \unitdb, thus creating an accessible dimensional knowledge system (\unitsys) to help handling tasks related to quantities.

Second, we propose \uniteval benchmark, aiming at both probing and enhancing the dimension perception skills of LLMs. The benchmark is designed to evaluate the capability of LLMs from three perspectives: \textit{basic perception}, \textit{dimension perception}, and \textit{scale perception}.
We proposed a bootstrapping strategy and a semi-automated approach to construct the datasets, aiming at reducing the costs associated with manual annotation.
We effectively enhance the model's capability to perceive dimensions by finetuning on these tasks.

Finally, we conduct experiments on both dimension perception and quantitative reasoning tasks.
Due to the lack of benchmarks tailored for quantitative reasoning, we extend numerical math word problems (N-MWP) to quantitative math word problems (Q-MWP).
By applying dimension perception techniques, such as quantity-oriented data augmentation, we showcase the enhancement brought by dimension perception on quantitative reasoning tasks.

\noindent \textbf{Contributions.} To summarize, we address the challenge of LLMs struggling with precisely understanding and reasoning about quantities in text. Our contributions are as follows:

\begin{itemize}
 \item  We present a framework that enhances quantitative reasoning by focusing on dimension perception. To the best of our knowledge, we are the first work to incorporate the concept of dimension into quantitative reasoning tasks.
 \item We construct a dimensional unit knowledge base (\unitdb) to supplement information regarding units, with a particular emphasis on dimensions.
 \item We propose \uniteval, a benchmark including seven tasks of three categories to probe and enhance the dimension perception skills of LLMs.
\item We systematically evaluate the performance of our model on quantitative reasoning tasks upon the enhancement through dimension perception.
The experiments demonstrate that our method boosts the performance of models with fewer parameters, like LLaMA-7B, raising the accuracy of quantitative reasoning from 43.55\% to 50.67\% compared to GPT-4.
\end{itemize}

\section{Preliminaries and framework}

\subsection{Preliminaries}

Table~\ref{tab:notations} summarizes the frequently used notations in this paper.

\begin{table}[h] 
    \centering
    \caption{Notations in this paper.}   
     \renewcommand{\arraystretch}{1.2}
    \label{tab:notations}
    \resizebox{0.37\textwidth}{18mm}{
        \begin{tabular}{ccc}
        \toprule
       \textbf{Symbol} & \textbf{Description} & \textbf{Example} \\
        \midrule 
        $v$  & Abstract Value & \textit{2} \\
        $u$ & Unit  & \textit{gill/h} \\
        $q$  &  Quantity & \textit{2 gill/h} \\
        $U$ & The unit Set  & \textit{\{meter, gill, ...\}}\\
        $K$ & QuantityKind & \textit{VilumeFlowRate} \\
        $dim(q)$ & Dimension of quantity $q$ & $L^3T^{-1}$\\
        $op$ & Arithmetic operation & $\times, \div$ \\
        $E$ & Arithmetic expression of units & $Joule \times Meter$ \\
        \bottomrule
        \end{tabular}
    }
\end{table}

\label{sec:preliminary}

\noindent \textbf{Dimension.}
Dimension is the fundamental attribute of quantities. 
It is the abbreviation for dimensional product or dimensional formula, essentially represented as the product of the powers of several base quantities to express derived physical quantities~\cite{international2001international}.
Each quantity $q$ can have its dimensional formula written as follows:
$$
dim(q)= L^\alpha M^\beta H^\gamma E^\sigma T^\epsilon A^\zeta I^\eta
$$
\noindent where the seven bases of dimensions are the fundamental physical quantities as shown in Table~\ref{tab:qudt-dim}, and $\alpha$, $\beta$, $\gamma$, $\sigma$, $\epsilon$, $\zeta$, $\eta$ are the dimensional exponents.
In the example shown in Fig.~\ref{fig:intro}, the dimension of quantity ``$0.1\ poundal$'' is $LMT^{-2}$.
\noindent \textbf{Quantitative Reasoning.}
Quantitative reasoning is a category of mathematical reasoning tasks, requiring extensive calculations among quantities, including computations among values and units.
In this paper we utilize quantitative math word problem (Q-MWP) to evaluate the performance on quantitative reasoning tasks, an example of which is shown in Fig.~\ref{fig:intro}.

\subsection{Method Overview}
The framework of this paper comprises three steps, as illustrated in Fig. \ref{fig:overview}.
\begin{enumerate}
    \item \textbf{\unitdb Construction.}
    In Fig.~\ref{fig:overview} (a), we construct a dimensional unit knowledge base (\unitdb), which stores basic information of the frequently used units of quantities and their corresponding dimensions.
    Based on the knowledge base, we implement a context-based unit linking module to link the units of quantities in texts to \unitdb.
    \unitdb serves as the foundation of our work. 
    \item \textbf{Dimension Perception.}
    In Fig.~\ref{fig:overview} (b), we propose \uniteval benchmark including seven tasks of three categories relevant to quantities and dimensions.
    We design various data construction methods, including semi-automation and bootstrapping retrieval for different tasks.
    Through finetuning on these tasks, we incorporate dimensional knowledge into our \dimperc model.
    \item \textbf{Quantitative Reasoning.}
    In Fig.~\ref{fig:overview} (c), we introduce quantitative reasoning tasks. We extend numerical math word problems to quantitative math word problems (Q-MWP).
    We utilize two methods of dimension perception to enhance quantitative reasoning: one based on the pretrained model \dimperc and the other employing quantity-oriented data augmentation methods.
\end{enumerate}
\section{Dimensional Knowledge System}

\label{sec:system}
In this section, we develop a dimensional knowledge system (\unitsys). 
\unitsys consists of a dimensional unit knowledge base (\unitdb) and a unit linking module.
\unitdb stores a comprehensive collection of units and their dimensional information (Section \ref{sec:system_kb}) and the unit linking module servers to link the units in texts to \unitdb (Section \ref{sec:system_link}). 

\subsection{Dimensional Unit Knowledge Base}
\label{sec:system_kb}

\subsubsection{Design principles}
Our design for \unitdb follows these principles:
\begin{itemize}
    \item \textit{Completeness}: \unitdb aims to offer a comprehensive collection of dimensional units that encompasses a wide range of significant ones.
    \item \textit{Diversity}: \unitdb contains not only unit names but also a diverse range of unit expressions and associated information.
    \item \textit{Scientificity}: The architecture of \unitdb is grounded upon well-established scientific standards and methodologies.
    \item \textit{Reliability}: \unitdb emphasizes the precision and trustworthiness of its data.
\end{itemize}

In the following sections, we will showcase how the methods we used to construct \unitdb align with these principles.

\subsubsection{Data sources}
The data of \unitdb mainly comes from the QUDT\footnote{The resources are available at https://qudt.org. QUDT is under a Creative Commons License
(CC BY 4.0, https://creativecommons.org/licenses/by/4.0/). 
Our usage for research aligns with the requirements under the license.}, an English database in the field of measurement units based on ontology, encompasses international unit usage standards such as SI, BIPM, ISO, NIST, and others. 
We extend the database schema by incorporating additional features such as unit frequency and keywords to enhance its \textit{comprehensiveness}.
Additionally, we utilize machine translation combined with manual corrections to extend the database to support both Chinese and English language and manually add some Chinese-specific units to cater to the Chinese context.
We refer to authoritative data sources, such as Wikipedia and Baidu Baike, to guide our manual verification and corrections.
To ensure the accuracy of our data, we adopt a multi-round cross-validation strategy. 
In each validation iteration, every data item undergoes examination by at least three reviewers. 
If any of them identify an inconsistency, a stringent verification process is initiated.
Authoritative data sources combined with a multi-round cross-validation mechanism ensure the \textit{Reliability} of \unitdb.

Table~\ref{tab:qudt_features} illustrates the schema of \unitdb.
In addition to the unit name, each unit is equipped with symbols and aliases as its representations, along with various metadata such as descriptions and keywords.
The diverse unit representations and the rich metadata support its \textit{Diversity}.

\begin{table}[h] 
    \centering
    \caption{Schema of Unit Data Stored in the Dimension Unit Knowledge Base (\unitdb)}   
     \renewcommand{\arraystretch}{1.2}
    \label{tab:qudt_features}
    \resizebox{0.40\textwidth}{23mm}{
        \begin{tabular}{c|c}
        \toprule
       \textbf{Feature}        & \textbf{Description} \\
        \midrule 
        UnitID &   Unit identifier           \\
        Label\_zh   &   Name of the unit in Chinese                \\
        Label\_en  &   Name of the unit in English           \\
        Symbol  &   Symbolic expressions of the unit          \\
        Alias  &   Alternative expressions of this unit in text          \\
        Description  &   A descriptive text for the unit           \\
        Keywords  &   Desciptive keywords for the unit           \\
        Frequency  &  The frequency of unit occurrence in texts        \\
        QuantityKind  &   The type of quantity represented by this unit    \\
        DimensionVec  &   The dimensional vector of this unit    \\
        ConversionVal  &   The converted value to the standard unit  \\
        \bottomrule
        \end{tabular}
    }
\end{table}

\subsubsection{Dimension Vector}

Each unit in \unitdb is associated with a \texttt{DimensionVec} feature that indicates the corresponding dimension type.
In line with \textit{Scientificity Principle}, we design the dimension vector based on the definition of dimension in physics.
In \unitdb, the fundamental units and symbols corresponding to each dimension are listed in Table~\ref{tab:qudt-dim}.

\begin{table}[h] 
    %
    
    \centering
    \caption{The symbol of eight dimensions and their corresponding fundamental quantities in DimensionVec feature.}   
    \label{tab:qudt-dim}
     \renewcommand{\arraystretch}{1.2}
    \resizebox{0.42\textwidth}{18mm}{
        \begin{tabular}{c|c|c|c}
        \toprule
       \textbf{Dim.}        & \textbf{Fundamental Quantity}  & \textbf{Basic Unit} & \textbf{Symbol}\\
        \midrule
        A &   \textbf{\textcolor{red}{A}}mount of Substance                              &  Mole      &      mol    \\
        E   &   \textbf{\textcolor{red}{E}}lectric Current                               &  Ampere    &      A       \\
        L  &   \textbf{\textcolor{red}{L}}ength                                          &  Metre     &      m       \\
        I  &   Luminous \textbf{\textcolor{red}{I}}ntensity                              &  Candela   &      cd      \\
        M  &   \textbf{\textcolor{red}{M}}ass                                            &  Kilogram  &      kg       \\
        H  &   T\textbf{\textcolor{red}{h}}ermodynamic Temperature                       & Kelvin     &      K        \\
        T  &   \textbf{\textcolor{red}{T}}ime                                            & Second     &      s        \\
        D  &   \textbf{\textcolor{red}{D}}imensionless                                   &  -         &      -        \\ 
        \bottomrule 
        \end{tabular}
    }

\end{table}

In the example depicted in Fig.~\ref{fig:overview} (a), ``\textit{dyne per centermeter}'' represents a unit of the quantity kind \textit{FourcePerLength}. Consequently, its dimensional expression corresponds to $MT^{-2}$, which is formally represented as ``\textit{A0E0L0I0M1H0T-2D0}'' in vector form.
The rules that dimensions obey are called the Dimension Laws, which play a significant role in quantitative reasoning. 
These laws assert that only physical quantities with identical dimensions can be added, subtracted, or compared.

\subsubsection{Unit Frequency}
Humans have a preference for utilizing certain units to express quantities, such as the frequent use of ``\textit{meters}'' and ``\textit{centimeters}'' and the rare use of ``\textit{decimeters}''.
To enhance the ability of language models to distinguish between frequently used and rare units, we incorporate the frequency attribute into all units in \unitdb, denoting their commonness in real-world scenarios.

We employ three criteria to evaluate the frequency of the unit: The degree of popularity measured by Googletrend\footnote{https://trends.google.com/trends/}; the assessment of a unit's commonality by human evaluators; and the frequency of the unit's occurrence in corpora.
Due to the challenge of identifying appropriate corpora with sufficient unit-rich content, we utilize tail entities from CN-DBpedia~\cite{xu2017cn} as a substitute for linguistic data. 
The frequency of unit occurrence in the general corpus is approximated by the frequency of unit occurrence in the tail entities, from which we derive the following formula for calculating unit frequency:
\begin{equation}
    Score(u) = \sum_{j \in \{GT,HS,CF\}} \alpha_j \cdot \log(Freq_j(u))
\end{equation}
\begin{equation}
    Freq(u) = (1-\delta)\cdot \frac{Score(u) - \min\limits_{u}{Score(u)}}{\max\limits_{u}{Score(u)} - \min\limits_{u}{Score(u)}} + \delta
\end{equation}
\noindent where $Freq_{GT}$ represents the frequency of the entity in Google Trend, $Freq_{HS}$ represents the human-scored frequency of the entity, and $Freq_{CF}$ represents the frequency of the entity in corpora. 
Additionally, $\alpha_{GT}$, $\alpha_{HS}$, $\alpha_{CF}$ and $\delta$ are weighting parameters that have been set to 0.3, 0.3, 0.4, and 0.1 in this paper.

\subsubsection{Statistics of \unitdb}
Table \ref{tab:qudt_statistic} provides a detailed comparison of \unitdb against the unit data used in UoM~\cite{park-etal-2022-language} and the mathematical engine \textit{WolframAlpha}\footnote{Wolfram Alpha, available at https://www.wolframalpha.com/, is a widely-used computational engine. Statistics from Wolfram Alpha are gathered via its API.}.
Remarkably, \unitdb encompasses a significantly greater number of units, spanning a broader variety of types.
An advantage of \unitdb is its bilingual support in both Chinese and English, while the other two databases only serve English-speaking users.
We also introduce the \texttt{DimensionVec} feature, which captures the intrinsic properties of each unit.
Furthermore, in contrast to \textit{WolframAlpha}'s closed-source interface, our database stands out by being fully open-source, transparent, and customizable.
Given these factors, our contribution offers a more thorough and linguistically inclusive approach to representing quantities in this domain.
Statistical data fully validates that \unitdb adheres to the \textit{Completeness Principle}.

\begin{figure}[t]
    \centering
    \includegraphics[width=0.49\textwidth, height=52mm]{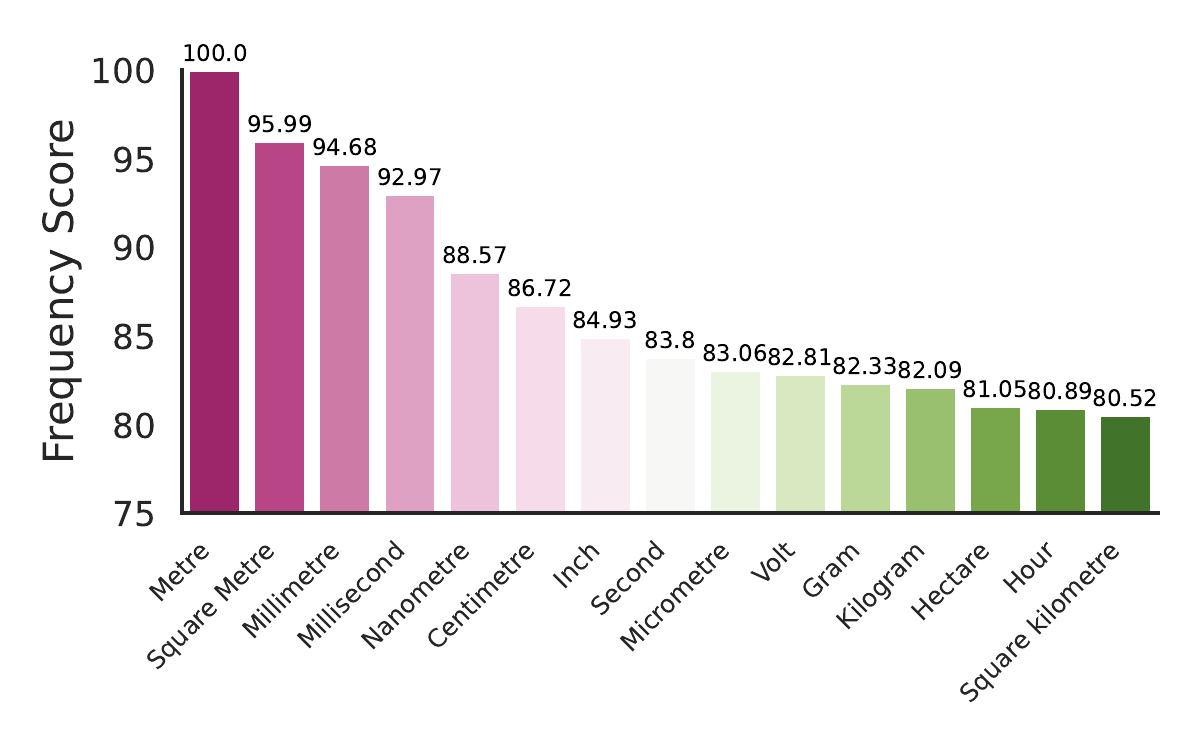}
    \caption{Popular units sorted by frequency feature in \unitdb.}
    \label{fig:top_frequency_unit}
\end{figure}

\begin{table}[ht]
    %
    
    \centering
    \caption{The statistics of DimUnitDB in comparison to UoM.}
    \label{tab:qudt_statistic}
     \renewcommand{\arraystretch}{1.2}
    \resizebox{0.48\textwidth}{8mm}{
        \begin{tabular}{c|c|c|c|c|c}
        \toprule
        \textbf{Resource} & \textbf{\# Units} &  \textbf{\# Quantity Kind}  &  \textbf{\# Dim. Vector} & \textbf{Lang.} & \textbf{Freq.}\\
        \midrule
        UoM\cite{park-etal-2022-language} & 76 & 16 & - & En & \ding{55} \\
        WolframAlpha & 540 & 173 & 63 & En & \ding{55} \\
        \textbf{DimUnitDB} & 1778 &  327  & 175 & En\&Zh & \ding{51} \\
        \bottomrule
        \end{tabular}
    }
     
\end{table}

\begin{figure}[t]
    \centering
    \includegraphics[width=0.40\textwidth]{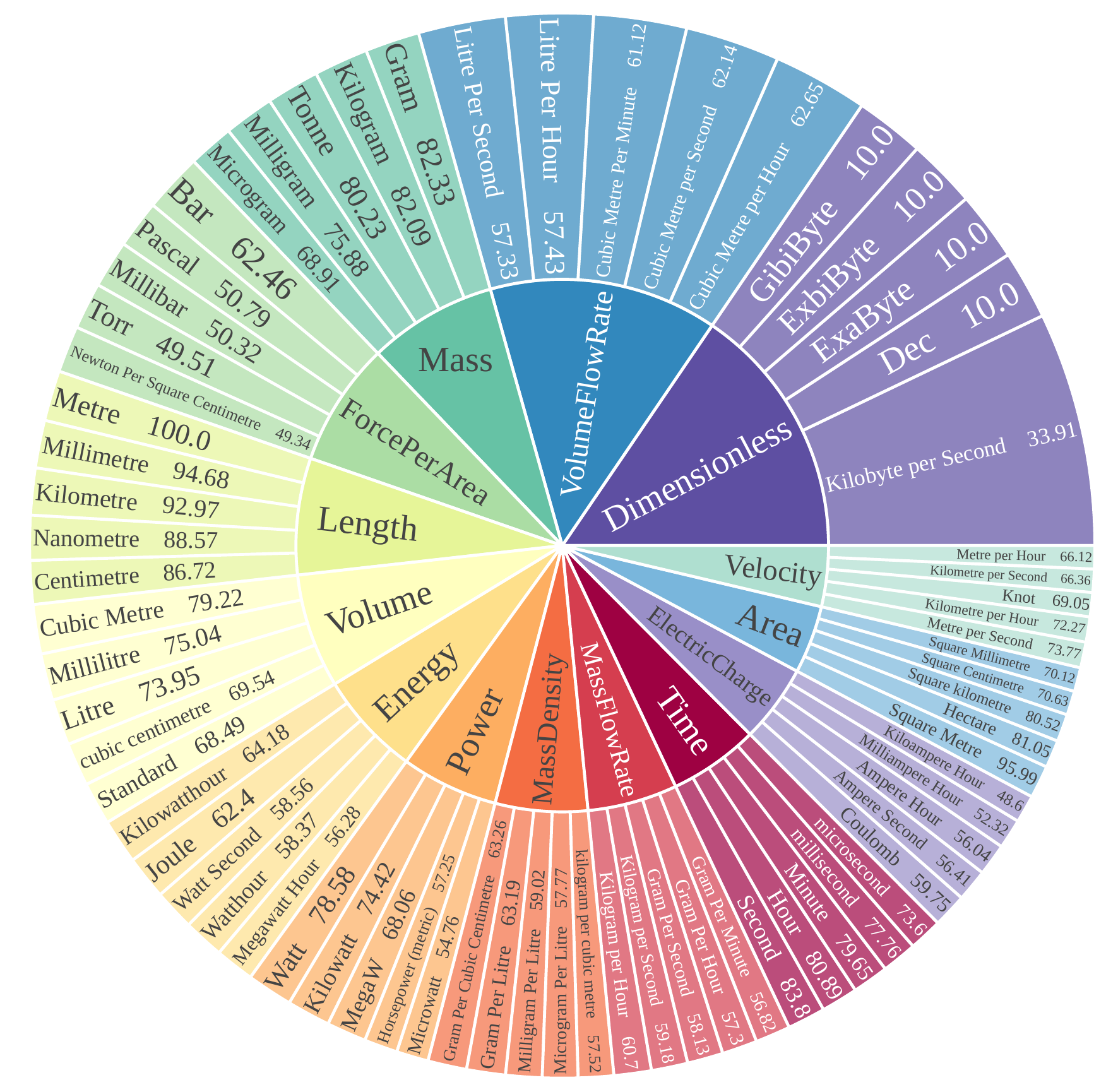}
    \caption{Top fourteen quantity kinds and their corresponding top five units, with the numerical values denoting their frequency feature in \unitdb.}
    \label{fig:qudt_sunburst}
\end{figure}

Units in \unitdb are sorted by frequency, as shown in Fig. \ref{fig:top_frequency_unit}. 
The frequency of \textit{QuantityKind} is determined by averaging the top five corresponding units, depicted in Fig. \ref{fig:qudt_sunburst}.

\subsection{Unit Linking Module}
\label{sec:system_link}

The depiction of units in natural language texts may exhibit irregularity.
A single unit may have multiple representations, and likewise, a single expression may indicate multiple units.
Similar to mapping entity mentions to knowledge graphs, we need to create a mapping from text mentions to \unitdb to accurately identify unit mentions.
For example, we link the unit mention ``dyne/cm'' from the question in Fig.~\ref{fig:intro} to ``DYN-PER-CentiM'' in \unitdb. 
The unit linking task can be formally defined as follows:
\begin{definition}[Unit Linking]
Given the contextual information $\mathbf{c}$ and a mention $\mathbf{m}$ of a unit within the text, map it to the corresponding dimensional unit $\mathbf{u} \in \unitdb$.
\end{definition}

To accomplish unit linking, the following steps are taken.

\subsubsection{Candidate unit generation} 
Key-values in the naming dictionary are retrieved and calculated the similarity with unit mentions to obtain all possible candidate units.
If the similarity exceeds a preset threshold, the corresponding unit will be added to the candidate list. 
We use the Levenshtein Distance as the similarity calculation metric, which can also be considered as the probability that a unit mention refers to a unit entity, that is, $Pr(\mathbf{u}|\mathbf{m})=Levenshtein Distance(\mathbf{u}, \mathbf{m})$.

\begin{figure*}[t]
    \centering
    \includegraphics[width=1\textwidth,height=38mm]{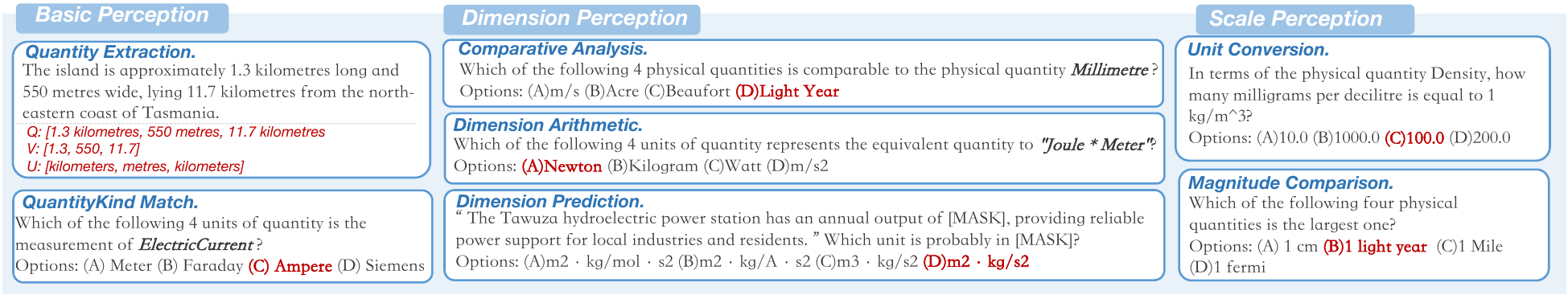}
    \caption{Illustrative examples of \uniteval. }
    \label{fig:example}
\end{figure*}

\subsubsection{Context-based Coreference Resolution}
We approximate the prior probability $Pr(\mathbf{u})$ of a unit in \unitdb by the frequency value calculated in Section \ref{sec:system_kb}, thus yielding $Pr(\mathbf{u}) = Freq(u)$.
Since a unit is usually mentioned in different contexts, the confidence of unit linking also relies on the contexts.
For instance,``\textit{degree}'' in different contexts might correspond to ``\textit{degrees Celsius}'' or ``\textit{diopter}''.
Specifically, given a context $\mathbf{c}$, we need to calculate the probability that it refers to the unit $\mathbf{u}$.
Initially, we apply the Word2Vec~\cite{mikolov2013efficient} tokenizer to segment the context $\mathbf{c}$ into ${c_i}$ and the stored keywords ${k_j}$ in \unitdb corresponding to $\textbf{u}$.
By calculating the cosine similarity between the segmented words and keywords, we approximate this probability by aggregating through similarity, as shown in the following formula:
$$
Pr(\mathbf{u}|\mathbf{c}) = \frac{1}{n} \sum_{i=1}^{n} \max_j \{CosSimilarity(c_i, k_j)\}
$$
\noindent where $n$ represents the number of contextual words. Therefore, the linked unit should be the one that offers the highest probability based on the given conditions of $\mathbf{c}$ and $\mathbf{m}$, denoted as $\tilde{u}$. 
Assuming that c and m are independent of each other, the calculation formula is:
$$
   \tilde{u} = \arg\max_{u} Pr(\mathbf{u}|\mathbf{c},\mathbf{m}) = \arg\max_{u} Pr(\mathbf{u})Pr(\mathbf{u}|\mathbf{m})Pr(\mathbf{u}|\mathbf{c})
$$

The unit linking module finally returns the result that all candidate units will be sorted in a descending order according to the confidence.

\section{Dimension Perception}

\label{sec:dimperc}
In this section, we propose \uniteval, a benchmark comprising seven tasks of three categories to evaluate the dimension perception skills of LLMs (Section~\ref{sec:three_levels},~\ref{sec:dimperc_task}).
We crawl abundant corpora from multiple sources and employ automated approaches to construct datasets (Section~\ref{sec:dimperc_dataset}).
We incorporate dimension knowledge and capabilities into the model through finetuning (Section~\ref{sec:dimperc_model}).

\subsection{Three Aspects of \uniteval}
\label{sec:three_levels}
We evaluate the capability of LLMs in understanding quantities from the following three aspects.

\subsubsection{Basic Perception}
The basic understanding of quantities involves identifying quantities and their components in texts, and matching them with the right quantity kinds.
We assess this through \textit{Quantity Extraction} and \textit{QuantityKind Match} tasks.

\subsubsection{Dimension Perception}
The concept of dimensions is the fundamental difference between quantities and abstract values.
Abstract values only retain one-dimensional scale, while the dimension attributes of units endow quantities with multi-dimensional characteristics.

In physics, there are several core premises concerning dimension. 
Firstly, the dimension law states that only quantities with the same dimension are comparable.
Secondly, units not only serve the function of representing magnitudes but also follow specific arithmetic rules. 
Lastly, there exists an inherent relationship between quantity kind and its corresponding dimension, allowing the prediction of dimension based on the natural language texts. 
Based on these principles, we design \textit{Comparable Analysis}, \textit{Dimension Arithmetic} and \textit{Dimension Prediction} to assess the dimension perception capabilities of LLMs.

\subsubsection{Scale Perception}
Different units within the same dimension signify varying magnitudes of measurement.
We design \textit{Unit Conversion} and \textit{Magnitude Comparison} tasks to assess the ability of LLMs to comprehend the scale of quantities.

\subsection{Task Definition}
We formalize the above-mentioned tasks as follows.
The notations used can be found in Table~\ref{tab:notations}.

\label{sec:dimperc_task}
\begin{definition}[Quantity Extraction]
Given a natural language text $X = \{x^1, x^2, ..., x^n\}$, the task of quantity extraction aims to generate a quantity list $Q = \{q^1, q^2, ..., q^k\}$, where each $q^i$ corresponds to $v^i$ and $u^i$, respectively representing the numerical and unit parts of the quantity.
\end{definition}

\begin{definition}[QuantityKind Match]
Given a specific quantity kind $K$ and a set of candidate units 
$U = \{u^1, u^2, ..., u^m\}$, the task is to select a correct unit $\hat{u}=u_j$ from $U$, where $\hat{u}$ describes the quantity kind $K$.
\end{definition}

\begin{definition}[Comparable Analysis]
Given two units \(u_1\) and \(u_2\), the task is to determine whether they are comparable, that is, whether they share the same dimension.
\end{definition}

\begin{definition}[Dimension Prediction]
Given a natural language text $X = \{x^1, x^2, ..., x^n\}$ in which certain quantities are represented by the [MASK] token, and a set of candidate units $U = \{u^1, u^2, ..., u^m\}$, the task is to select a unit $\hat{u} \in U$ such that $dim({\hat{u}})$ is consistent with the dimension implied by the [MASK] token in the context of $X$. 
\end{definition}

\begin{definition}[Dimension Arithmetic Task]
Given an arithmetic expression $E = u^1 op^1 u^2...op^{n-1} u^n$ and a set of candidate units $U = \{u^1, u^2, ..., u^m\}$, the task is to predict a unit $\hat{u} \in U$ such that $dim({\hat{u}})$ = $dim(E)$.
\end{definition}

\begin{definition}[Magnitude Comparison]
Given a set of candidate units \( U = \{u^1, u^2, ..., u^m\} \), the task of magnitude comparison requires the identification of the unit $\hat{u} \in U$such that $\hat{u}$ possesses the largest magnitude among all units in $U$. 
\end{definition}

\begin{definition}[Unit Conversion]
Given two units $u_1$ and $u_2$, the task is to determine a real number $\beta \in \mathbb{R}$, such that $u_1$ can be converted to $u_2$ by multiplying with $\beta$. Formally, $u_1 \times \beta = u_2$.
\end{definition}

We integrate these seven tasks into \uniteval benchmark, to evaluate whether a model possesses sufficient dimension perception capability.
Since binary classification can be easily generalized to multi-way choice, we convert all judgment tasks into selection tasks and present illustrative examples in Fig.~\ref{fig:example} of \uniteval.
In all examples, the number $m$ of candidate units is set to 4.

\subsection{Dataset Construction}
\label{sec:dimperc_dataset}

Among the seven tasks mentioned in Section \ref{sec:dimperc_task}, both quantity extraction and dimension prediction require the representation of quantities in natural language texts, thus necessitating the specialized construction of their corresponding datasets. 
The remaining five tasks can be constructed with the corresponding datasets through the heuristic rule-based methods with \unitsys.
We present the method design for constructing the quantity extraction and dimension prediction datasets in Section~\ref{sec:qe_dataset} and Section~\ref{sec:dp_dataset}, respectively.

\begin{algorithm}[t]
    \footnotesize
    \SetAlgoLined
    \caption{Semi-automated Annotating Method}
    \label{algorithm:semi_label}
    \KwData{Corpus $C$, \unitsys annotator $D$, PLM $M$}
    \KwResult{Annotated Dataset $D'$}
    Initialize Annotated Corpus $C_0 \leftarrow \emptyset$\;
    \tcp{Step 1: Initially annotate with DimKS}
    \For {each sentence $s$ in $C$}{
        $text \gets$ preprocessed version of $s$\;
        $s_1, labels \leftarrow D.annotate(s)$\;
        \If {($s_1$ contains numeric entity)}{
            Append $(s_1, \text{labels})$ to $C_0$\;
        }
    }
    \tcp{Step 2: Retrieve and filter with PLM}
    Initialize Annotated Dataset $D' \leftarrow \emptyset$\;
    \For {each tuple $(s, \text{labels})$ in $C_0$}{
        $s_{\text{masked}} \leftarrow$ Replace numerics in $s$ with [MASK]\;
        $p \leftarrow M.predict(s_{\text{masked}})$\;
        \If {($p$ is numeric)}{
            Append $(s, \text{labels})$ to $D'$
        }
    }
    \tcp{Step 3: Manual Review and Refinement}
    Review and refine $D'$ manually, finalize $D'$\;
    \Return $D'$
\end{algorithm}

\subsubsection{Semi-automated Annotating}
\label{sec:qe_dataset}
We first crawl corpora abundant in quantities from sources including high-school physics test websites, electronic information forums, industrial knowledge graphs, and a general domain knowledge graph~\cite{xu2017cn}.

A semi-automated method is then proposed to label the corpus, as depicted in the Algorithm \ref{algorithm:semi_label}.
A heuristic rule-based approach is employed for initial annotation, utilizing regular expressions to extract values, followed by attempts to link subsequent mentions into \unitdb as units.
If unit linking is successful, the position of the quantity mention is annotated.

However, the heuristic rule-based method might mislabel expressions that don't signify quantities, such as ``LPUI-1T'', a device code, by wrongly interpreting the ``1T'' phrase as ``1 ton'' or ``1 Tesla''.
To mitigate this issue, we utilize a large-scale pretrained language model for error detection. 
This is done by substituting quantity mentions in the text with a ``[MASK]'' token, followed by leveraging BERT~\cite{devlin-etal-2019-bert} to infer the masked words.
If the predicted words cannot be linked with \unitdb, the reference is considered non-quantitative and removed through filtering.
Our approach achieves an annotation accuracy of 82\%. 
Ultimately, manual inspection is required to rectify the labeling inaccuracies.

\subsubsection{Bootstrapping Retrieval}
\label{sec:dp_dataset}
The dataset of dimension prediction is derived from CN-DBpedia~\cite{xu2017cn}, given its semi-structured format and plentiful \texttt{<subject, predicate, object>} triplets.
We extract specific triplets, like \texttt{<LeBron James, height, 2.06 meters>}, where predicate and object denote quantitykind and value, using the bootstrapping method outlined in Algorithm~\ref{algorithm:bootstrap}.

We maintain a dynamic mention set and a predicate set representing quantities. 
Initially, we fetch high-frequency units from \unitdb as the mention set. 
A three-step bootstrapping method is applied to the two sets. 
The first step updates the predicate set by leveraging the mention set and graph search. 
The second step involves filtering the predicate set, where we calculate the proportion of quantity-related triplets and retain those predicates that meet a certain threshold. 
In the third step, the mention set is expanded using the predicate set and graph search. 
After several iterations, we eventually retrieve quantitative triplets by the mention set and predicate set.

To better align with natural language, we feed the quantity triplets into ChatGPT to generate sentences that include these triplets and annotate the resulting dataset with \unitsys.
Quantities in texts are masked and annotated by their corresponding dimension, forming the dimension prediction dataset.

\subsubsection{Complexity Analysis}

In Algorithm~\ref{algorithm:semi_label}, automated annotation includes the first two steps. In Step 1, we use \unitsys to annotate the corpus. Let $|C|$ and $l$ be the scale of the corpus and the average length of sentences, the time complexity is $O(l|C|)$. In Step 2, we use a PLM to filter the annotated corpus $C_0$, the upper limit of the scale of $C_0$ is the scale of $C$. Let $k$ be the time that the PLM predicts a text, the time complexity is $O(k|C|)$. The total time complexity is $O(l|C|+k|C|)$.

Algorithm~\ref{algorithm:bootstrap} includes a bootstrapping process with three steps. Throughout this process, we continuously maintain the growing mention set $M$ and predicate set $P$ and finally obtain the quantitative triplets set $T$. Let $|M|$, $|P|$ and $|T|$ be the average length of these sets during the process and retrieval time in CN-DBpedia be $r$. The time spent per iteration is $O(r|M|+2r|P|+r|T|)$, so the total time spent for $\delta$ iterations is $O(\delta(r|M|+2r|P|+r|T|))$. In our setup, $\delta$ is 5, so our algorithm is efficient in terms of the time complexity.

\subsection{Dimensional Perception Methodology}
\label{sec:dimperc_model}
We employ the dimension perception tasks outlined in Section \ref{sec:dimperc_task} as training tasks, enabling the continuous finetuning of the language model. 
This infuses dimensional knowledge into the language model, concurrently enhancing its capacity for dimensional perception.

\noindent\textbf{Reasoning Process.} 
To improve the precision of the model's responses, we expect the model's output to express a detailed process of thought, followed by reasoning.
While constructing the ground truth answers, we utilize templates to generate corresponding Chain of Thought (CoT)~\cite{wei2022chain} reasoning procedure to structure the output.
We denote the sequence of the reasoning process as $R$ and the sequence of answering the question as $A$, therefore, the output sequence can be expressed as $\mathbf{y}=$``$\texttt{<bos>}\ R\ \texttt{<sep>}\ A \ \texttt{<eos>}$".
The reasoning process of our training set is generated based on rules and templates.

\begin{algorithm}[t]
    \footnotesize
    \SetAlgoLined
    \caption{Bootstrapping Retrieval Method}
    \label{algorithm:bootstrap}
    \KwData{\unitdb, \unitsys, CN-DBpedia $K$, threshold $\tau$, bootstrapping iterations $\delta$}
    \KwResult{Quantitative triplets}
    Mention Set $M_0 \leftarrow$ highFreqUnits(\unitdb)\;
    \tcp{Bootstrapping}
    \For {each iteration $\delta$}{
        \tcp{Step 1: Initialize predicates set}
        $P \leftarrow \emptyset$\;
        \For {each mention $m$ in $M$}{
            $triplets \leftarrow$ findTriplets($K$, $m$ in object)\;
            Extract predicates and append them to $P$;
        }
        \tcp{Step 2: Filter predicates by ratio}
        \For {each predicate $p$ in $P$}{
            $triplets \leftarrow$ findTriplets($K$, $p$)\;
            $ratio \leftarrow$ calculateQuantityRatio($triplets$, \unitsys)\;
            \If {($ratio$ $<$ $\tau$)}{
                Remove $p$ from $P$;
            }
        }
        \tcp{Step 3: Update objectives set}
        $M \leftarrow \emptyset$\;
        \For {each predicate $p$ in $P$}{
            $triplets \leftarrow$ findTriplets($K$, $p$)\;
            Extract units mentioned in the objects, append them to $M$\;
        }
    }
    \tcp{Obtain final triplets}
    $quantitativeTriplets \leftarrow$ retrieveTriplets($K$, $P$, $M$)\;
    \Return $quantitativeTriplets$
\end{algorithm}

\noindent \textbf{Model.} Our methodology employs a standard Transformer model architecture, which operates solely on decoder-based attention mechanisms.
Initially, we encode the context and questions as input $\mathbf{x} = [x_1, ..., x_n]$. The objective is to minimize the negative log-likelihood of the target sequence $\mathbf{y}$ given the input context, which can be formally expressed as:
\begin{equation}
L(\theta) = \log p(\mathbf{y}|\mathbf{x};\theta) = -\sum_{i=1}^{m} \log P(y_i | y_{<i}, \mathbf{x}; \theta)
\end{equation}
\noindent where $\theta$ represents the parameters of the model.

\section{Quantitative Reasoning}
\label{sec:quanreason}

\begin{CJK}{UTF8}{gbsn}
\begin{table*}[!t]
    \centering
    \caption{Quantity-Oriented Augmentation Examples from all Proposed Methods.We have colored the substituted portions for visual distinction.}
    \label{tab:dataaug_example}
    \resizebox{0.99\textwidth}{26mm} {
        \begin{tabular}{clll}
        \toprule
                \multicolumn{2}{c}{\multirow{3}{*}{\textbf{Augmentation Method}}} & \multicolumn{2}{l}{\textbf{Examples}} \\
        \cmidrule{3-4}
               & & \textbf{Original} & 小王要将\textit{\textcolor{brown}{150千克}}含药量20\%的农药稀释成含药量5\%的药水。需要加水多少\textit{\textcolor{brown}{千克}}？\\

               & & & {\scriptsize Xiao Wang wants to dilute the \textit{\textcolor{brown}{150 kilograms}} pesticide containing 20\% into a potion containing 5\%. How many \textit{\textcolor{brown}{kilograms}} of water need to be added?} \\
        \midrule
               \multirow{6}{*}{\rotatebox{90}{\textbf{Context-based}}} & \textbf{Format Substitution} & \textbf{Augmented} & 
               小王要将\textit{\textcolor{brown}{150 kg}}含药量20\%的农药稀释成含药量5\%的药水．需要加水多少千克？\\
               
               & & & {\scriptsize Xiao Wang wants to dilute the \textit{\textcolor{brown}{150kg}} pesticide containing 20\% into a potion containing 5\%. How many kilograms of water need to be added?} \\

               & & \textbf{Answer} &
               450 $\rightarrow$ 450\\
        \cmidrule{2-4}
                & \textbf{Dimension Substitution} & \textbf{Augmented} & 
                小王要将\textit{\textcolor{brown}{150000克}}含药量20\%的农药稀释成含药量5\%的药水．需要加水多少千克？\\
                
                & & & {\scriptsize Xiao Wang wants to dilute the \textit{\textcolor{brown}{150000 grams}} pesticide containing 20\% into a potion containing 5\%. How many kilograms of water need to be added?} \\

               & & \textbf{Answer} & 
               450 $\rightarrow$ 450\\
        \midrule
               \multirow{6}{*}{\rotatebox{90}{\textbf{Question-based}}} & \textbf{Format Substitution} &  \textbf{Augmented} &
               小王要将150千克含药量20\%的农药稀释成含药量5\%的药水．需要加水多少\textit{\textcolor{brown}{kg}}？\\
               
                & & & {\scriptsize Xiao Wang wants to dilute the 150 kilograms pesticide containing 20\% into a potion containing 5\%. How many \textit{\textcolor{brown}{kgs}} of water need to be added?} \\
               
               & & \textbf{Answer} &
               450 $\rightarrow$ 450\\
        \cmidrule{2-4}
                & \textbf{Dimension Substitution} & \textbf{Augmented} &
                小王要将150千克含药量20\%的农药稀释成含药量5\%的药水．需要加水多少\textit{\textcolor{brown}{吨}}？\\

              & & & {\scriptsize Xiao Wang wants to dilute 150 kilograms of pesticide with a concentration of 20\% to a solution with a concentration of 5\%. How many \textit{\textcolor{brown}{tons}} of water need to be added?} \\

               & & \textbf{Answer}  & 
               450 $\rightarrow$ \textcolor{red}{\textbf{0.45}}\\
        \bottomrule
        \end{tabular}
    }
\end{table*}
\end{CJK}

In this section, we apply dimension perception to quantitative reasoning tasks. 
We first introduce the Quantitative Math Word Problem task (Section \ref{sec:qmwp}) and propose solutions for quantitative reasoning problems (Section \ref{sec:qmwp_method}).

\subsection{Quantitative Math Word Problem}
\label{sec:qmwp}

Math Word Problem (MWP) is a type of Machine Reading Comprehension (MRC) problem, where the text includes numerous values and presents real-world situations that need mathematical solutions.
Solving these problems requires both text comprehension and mathematical skills to extract relevant information and formulate appropriate equations.

However, existing MWP datasets~\cite{wang-etal-2017-deep,zhao2020ape210k} frequently exhibit uniformity in unit representation, showcasing insufficient diversity in dimensions, thus posing limited demands on the quantitative understanding of language models. 
In the context of this paper, we refer to these MWPs as Numerical Math Word Problems (N-MWPs).
Evaluating quantitative comprehension of LLMs with N-MWP datasets is inadequate, as models might only grasp numerical values without fully understanding units bearing dimensional concepts.

To address this, we innovatively propose a new task: Quantitative-Math Word Problem (Q-MWP) based on N-MWP.
The problem descriptions in Q-MWP feature a wide range of quantities with diverse dimensions, requiring models to not only possess numerical understanding and reasoning capabilities but also to have a comprehensive and robust cognition of quantities, particularly in the conceptual understanding of dimensions.
The format of Q-MWP is in alignment with regular MWP. 
Due to the variety of unit representations for the same dimension in Q-MWP, the reasoning process often requires the normalization or unification of units.

\subsection{Quantitative Reasoning with Dimension Perception}
\label{sec:qmwp_method}

\subsubsection{\dimperc as base model}
Section~\ref{sec:dimperc} utilizes seven tasks to incorporate dimensional knowledge and skills into language models.
We use \dimperc, derived from this process, as the base model for quantitative reasoning tasks. 
A comprehensive understanding of dimensions assists the model in reasoning about quantities effectively.

\subsubsection{Quantity-Oriented Data Augment}
\label{sec:qmwp_da}

The application of data augmentation methods to MWPs is recently confined to general text augmentation techniques or those concentrating specifically on numerical portions\cite{kumar2022practice,sivakumar2023fermat,zhou2023learning}. 
However, as units are crucial to quantities and the solving of mathematical word problems, we employ quantity-oriented data augmentation methods to boost the model's performance on Q-MWP problems. 
Using the \unitdb from Section \ref{sec:system}, we propose several data augmentation methods. 
Through these methods, we incorporate knowledge about units and dimensions into our model, improving its ability to understand quantities.

As shown in Table \ref{tab:dataaug_example}, we apply two major data augmentation directions: context-based data augmentation and question-based data augmentation. 
For each augmentation direction, we employ two substitute methods: \textit{Unit Format Substitution} and \textit{Substitution of Units with Same Dimension}.
Context-based method substitutes the contextual information in mathematical problems while ensuring the core meaning stays consistent.
It requires maintaining the invariant scale of quantities. 
This approach allows data augmentation while keeping the answer unchanged.
Question-based method substitutes the questioning part of mathematical problems. 
As this method alters the answer, simultaneous adjustments to the solution equation and answer are necessary for proper handling.
The \textit{Unit Format Substitution} method, in the context of unit replacement, preserves the unit itself and replaces the original representation with an equivalent alternative unit format. 
On the other hand, the \textit{Substitution of Units with Same Dimension} method replaces the original unit with a unit of the same dimension. 
In the context-based approach, both the numerical value and unit components of quantity terms need to be transformed simultaneously to maintain invariance in scale and size.

\subsubsection{Equation Tokenization}
Solving mathematical word problems requires the model to be sensitive to numbers. 
However, the work~\cite{thawani-etal-2021-representing} argues that general text-based methods are not suitable for handling numerical information.
Therefore, we employ a method similar to the digit tokenization technique proposed in~\cite{ggb2020injecting} to investigate its effectiveness in the context of LLMs and our specific task scenario.
Specifically, for a word-piece of an equation 
\texttt{\#\#$e_1,\cdots$,\#\#$e_k$} 
where \texttt{$e_i \in D \cup Op$}, \texttt{$D = \{0,...9\}$},
\texttt{$Op = \{+,-,*,/,\%,=,(,)  \} $ } 
is further split into \texttt{\#\#$e_1,\cdots,$\#\#$e_k$}.

\subsubsection{Training Detail}
We employ the Seq2Seq method for training, where during decoding, we first generate the solution equation and then provide the corresponding answer.
We denote the sequence of the inference equation as $E$ and the sequence of answering the question as $A$, therefore, the output sequence can be expressed as $\mathbf{y}=$``$\texttt{<bos>}\ E\ \texttt{<sep>}\ A \ \texttt{<eos>}$".

\section{Experiments}

\label{sec:exp}
In this section, we conduct experiments on dimension perception and quantitative reasoning tasks to answer the following research questions:
\begin{itemize}
  \item[\textbf{$\bullet$}] \textbf{RQ1}: Do LLMs have sufficient knowledge of dimension perception of quantities?
  \item[\textbf{$\bullet$}] \textbf{RQ2}: Is our continual fintuning approach for dimension perception (Section~\ref{sec:dimperc}) effective to improve the corresponding capability of LLM?
  \item[\textbf{$\bullet$}] \textbf{RQ3}: Are current LLMs able to do reasoning tasks dense with quantities? Do our methodologies (Section~\ref{sec:qmwp_method}) improve the reasoning capabilities of existing LLMs?
 \item[\textbf{$\bullet$}] \textbf{RQ4}: Compared to those methods that enhance LLM by external tools, are there any advantages of our approach?

\end{itemize}

\subsection{Experiment Datasets}

\subsubsection{Dataset of Dimension Perception} 
We utilize the \uniteval benchmark in Section~\ref{sec:dimperc} to assess whether language models possess sufficient knowledge to support their dimension perception of quantities.
This benchmark includes seven sub-tasks of three categories, thoroughly assessing language models in basic perception, dimension perception, and scale perception.

\subsubsection{Dataset of Quantitative Reasoning} 
\label{sec:qmwp_data}

We employ two fundamental N-MWP datasets, Math23k~\cite{wang-etal-2017-deep} and Ape210k~\cite{zhao2020ape210k}, which include Chinese elementary mathematical problems.
These datasets are larger and more complex than GSM8k, necessitating advanced mathematical understanding due to multi-step calculations.
We select these two datasets for their alignment with \unitdb, yet the techniques outlined in this paper are adaptable and applicable to any language framework.
For testing convenience, we extract subsets from Math23k and Ape210k for evaluation, denoted as N-Math23k and N-Ape210k respectively.
Similar to the methodology in Section \ref{sec:qmwp_da}, we augment the above two datasets for quantitative richness, resulting in the expanded versions termed \textbf{Q-Math23k} and \textbf{Q-Ape210k}.

Due to the involvement of unit conversions in Q-MWP, they require more computational steps compared to N-MWP problems.
The statistics of the evaluation datasets are displayed in Table~\ref{tab:qmwp_dataset}.

\begin{table}[H]
    \centering
    \caption{Statistics of Evaluation Datasets on Quantitative Reasoning. 
    }
    \label{tab:qmwp_dataset}
    \resizebox{0.45\textwidth}{13mm}{
        \begin{tabular}{ccccccc}
        \toprule
        \multirow{2}{*}{Dataset} & \multirow{2}{*}{\#Num} & \multirow{2}{*}{\#Units} & \multicolumn{4}{c}{ \#Operations } \\
        \cmidrule{4-7} 
           &  &  &  [0,3] & (3,5] & (5,8] & (8,$+\infty$)\\
        \midrule
        N-Math23k &  225 & 17 & 162 & 47 & 16 & 0 \\
        N-Ape210k & 225 & 18 & 139 & 55 & 27 & 4 \\
        Q-Math23k &  225 & 35 & 108 & 86 & 24 & 7\\
        Q-Ape210k &  225 & 52 & 99 & 68 & 39 & 19 \\
        \bottomrule
        \end{tabular}
    }
\end{table}

\begin{table*}[t]

    \centering
    \caption{Results(\%) of different models and settings on \uniteval.}
    \label{tab:Dimperc_result}
    \renewcommand{\arraystretch}{1.25}
    
\resizebox{1\textwidth}{35mm}{
    \begin{tabular}{lcccccccccccccccc}
    \toprule
     &  & \multicolumn{5}{c}{\textbf{Basic Perception}}     & \multicolumn{6}{c}{\textbf{Dimension Perception}} & \multicolumn{4}{c}{\textbf{Scale Perception}} \\
    \cmidrule(lr){3-7} \cmidrule(lr){8-13} \cmidrule(lr){14-17} 
    \textbf{Model} & \textbf{\#params} & \multicolumn{3}{c}{\textbf{Quantity Extraction}}   & \multicolumn{2}{c}{\textbf{QuanKind Match}}  & \multicolumn{2}{c}{\textbf{Comparable Analysis}}   & \multicolumn{2}{c}{\textbf{Dimension Pred.}} & \multicolumn{2}{c}{\textbf{Dimension Arith.}}  & \multicolumn{2}{c}{\textbf{Magnitude Comp.}} & \multicolumn{2}{c}{\textbf{Unit Conversion}}  \\
    \cmidrule(lr){3-5} \cmidrule(lr){6-7} \cmidrule(lr){8-9} \cmidrule(lr){10-11} \cmidrule(lr){12-13} \cmidrule(lr){14-15} \cmidrule(lr){16-17}
      &  & \textbf{QE}   & \textbf{VE}  & \textbf{UE}  & \textbf{Prec.}  & \textbf{F1}  & \textbf{Prec.}   & \textbf{F1}  & \textbf{Prec.}  & \textbf{F1}  & \textbf{Prec.} & \textbf{F1}   & \textbf{Prec.}  & \textbf{F1}  & \textbf{Prec.}  & \textbf{F1}   \\

\midrule
\rowcolor[gray]{0.95} \multicolumn{17}{c}{{\textit{LLMs with external tool (w/ Wolfram Alpha)}}} \\ 
                           \multicolumn{1}{l}{GPT-4} & -
& 68.40 & 79.70 & 78.22 & 64.44 & 54.37 & 71.11 & 58.71 & 62.22 & 56.48 & 26.67 & 25.61 & 64.44 & 53.76 & 73.33 & 59.30 \\
    
                           \multicolumn{1}{l}{GPT-3.5-Turbo} & -
& 44.09 & 46.74 & 55.94 & 33.33 & 32.40 & 31.11 & 33.39 & 48.89 & 45.43 & 8.89 & 9.31 & 20.00 & 18.77 & 28.89 & 27.83 \\

\midrule
\rowcolor[gray]{0.95} \multicolumn{17}{c}{{\textit{powerful closed-source LLMs \& open-source LLMs ($>50B$)}}} \\ 

                           \multicolumn{1}{l}{GPT-4} & -
& {73.91} & \textbf{80.59} & 80.79 & \textbf{66.67} & 39.63 & 68.89 & 55.18 & 44.44 & 34.40 & 31.11 & 14.98 & 53.33 & 31.37 & 64.45 & 52.68 \\

                         \multicolumn{1}{l}{GPT-3.5-Turbo} & - 
& {73.48} & {78.18} & {78.95} & 46.00 & 18.43 & 39.91 & 24.63 & 47.56 & 25.05 & 19.50 & 7.38 & 39.73 & 13.71 & 41.96 & 23.42   \\

 \multicolumn{1}{l}{InstructGPT} & 175B 
& \textbf{77.67} & {76.57} & 80.70 & 49.50 & 32.99 & 42.15 & {42.42} & 54.47  & 43.24 &  24.00 & 15.70 & 37.50 & 28.12 & 60.71 & 59.80   \\

                         \multicolumn{1}{l}{PaLM-2} & 540B
& - & - & - & 68.89 & 47.29 & 51.11 & 44.67 & 53.33 & 31.24 & 31.11 & 23.11 & 17.78 & 15.65 & 60.00 & 38.90 \\

                        \multicolumn{1}{l}{LLaMa-2} & 70B
& 65.94 & 60.45 & 71.79 & 28.89 & 27.03 & 33.33 & 31.93 & 42.22 & 41.08 & 22.22 & 20.41 & 31.11 & 28.11 & 46.67 & 33.60 \\




\midrule
\rowcolor[gray]{0.95} \multicolumn{17}{c}{{\textit{open-source LLMs (1-50B)}}}\\ 
                        \multicolumn{1}{l}{LLaMa-2} & 13B
& 57.58 & 59.09 & 58.42 & 44.44 & 39.82 & 24.44 & 25.92 & 51.11 & 36.62 & 20.00 & 19.92 & 13.34 & 5.6 & 33.33 & 21.90 \\
                        \multicolumn{1}{l}{OpenChat} & 13B
& 33.07 & 39.69 & 46.23 & 37.77 & 30.33 & 28.89 & 22.01 & 35.56 & 26.75 & 26.67 & 20.84 & 20.00 & 14.17 & 28.89 & 24.26 \\
                       \multicolumn{1}{l}{Flan-T5}  & 11B      
& - & - & - & 40.00 & 36.00 & 37.78 & 32.15 & 47.11 & 39.67 & 17.00 & 14.95 & 16.07 & 15.49 & 30.80 & 23.27  \\
                          \multicolumn{1}{l}{T0++} & 11B 
& - & - & - & 18.76 & 17.26  & 18.67 & 17.26 & 41.33 & 36.88  & 6.00 & 6.99 & 15.62 & 16.74 & 13.39 & 17.20   \\
                        \multicolumn{1}{l}{ChatGLM-2} & 6B
& 36.30 & 35.29 & 45.25 & 44.44 & 34.89 & 42.22 & 32.71 & 28.89 & 25.15 & 17.78 & 14.77 & 20.00 & 18.45 & 24.44 & 19.93 \\
                         \multicolumn{1}{l}{\textbf{Dimperc (Ours)}} & 7B       
& 71.53 & 73.61 & \textbf{82.35} & {62.81} & \textbf{62.59} & \textbf{83.03} & \textbf{66.50} & \textbf{99.11} & \textbf{99.13} & \textbf{66.33} & \textbf{66.28} & \textbf{83.93} & \textbf{67.22} & \textbf{95.54} & \textbf{95.39} \\
        
    \bottomrule
\end{tabular}
}

\end{table*}
\begin{table}[t]
    \centering
    \caption{Comparison between \dimperc and the base model on \uniteval.}
    \label{tab:dimperc_ablation}
    \resizebox{0.48\textwidth}{9mm}{
        \begin{tabular}{ccccccc}
        \toprule
        \multirow{2}{*}{Model} & \multicolumn{2}{c}{Basic Perception} &  \multicolumn{2}{c}{Dimension Perception} & \multicolumn{2}{c}{Scale Perception} \\
        \cmidrule(lr){2-3}  \cmidrule(lr){4-5} \cmidrule(lr){6-7}
                  &  Prec. & F1      &  Prec. & F1    & Prec. & F1 \\
        \midrule
            \textbf{LLaMa\textsubscript{IFT}} & 29.65 & 24.01 & 20.38 & 16.64 & 8.94 & 6.70   \\
            \textbf{Dimperc}   & \textbf{71.69} & \textbf{63.13} & \textbf{82.82} & \textbf{77.30} & \textbf{89.74} & \textbf{81.31}     \\
        \bottomrule
        \end{tabular}
    }
\end{table}

\subsection{Baselines}
We compare our model with several large language models (LLMs) that have been trained on extensive corpora and are imbued with the ability to comprehend instructions, equipping them to tackle our dimension perception and quantitative reasoning tasks in either zero-shot or few-shot scenarios. 

The chosen LLMs include GPT-4~\cite{openai2023GPT4}, GPT-3.5-Turbo\footnote{https://chat.openai.com/chat/}, InstructGPT~\cite{ouyang2022training}, PaLM-2~\cite{anil2023palm}, LLaMa-2~\cite{touvron2023llama}, OpenChat~\cite{wang2023openchat}, Flan-T5~\cite{brown2020language}, T0++~\cite{sanh2021multitask}, ChatGLM-2~\footnote{https://github.com/THUDM/ChatGLM2-6B}.
In addition, in quantitative reasoning tasks, we compared our approach with specific models, BertGen~\cite{devlin-etal-2019-bert} and LLaMa~\cite{touvron2023llama}, which has been supervised trained on N-MWP datasets.

Our baselines also include tool-augmented LLMs. 
GPT-4 and GPT-3.5-Turbo have demonstrated their abilities to interface with external tools, so we employ them as the backbones for the tool-augmented baselines.
We use \textit{LangChain}~\footnote{https://www.langchain.com/} to facilitate the integration of the LLMs with the mathematics engine \textit{WolframAlpha}.

Due to the specific format required for quantity extraction tasks, we employ a few-shot scenario for testing, while all other tasks are assessed in a zero-shot scenario. 
Additionally, given the strong language dependency of extraction task and the lack of Chinese support in the PaLM-2's API, Flan-T5 and T0++, we are unable to evaluate their performance in quantity extraction. 
However, assuming a lower language dependency for other dimension perception tasks, we test the remaining tasks by translating the questions into English.

\subsection{Setting Details}
All of our experiments are conducted on the workstations of NVIDIA A800 PCIe with 80GB memory and the environment of Ubuntu 20.04.6 LTS and torch 2.0.1.

We first finetune LLaMA-7B~\cite{touvron2023llama} on a generic instruction dataset to equip the model with a foundational understanding of the tasks, named LLaMA\_IFT.
We employ LLaMA\_IFT as the base model in dimension perception tasks and use the model \dimperc, which is incorporated with dimension perception ability in Section~\ref{sec:dimperc}, as the base model in quantitative reasoning tasks.

\subsection{Metrics}
In the quantity extraction task, we apply F1-scores for evaluation, identifying the three sub-tasks as \texttt{QE}, \texttt{VE}, and \texttt{UE}.
We assess the other six dimension perception tasks with \texttt{Precision} and \texttt{F1-score} metrics.
For quantitative reasoning, we simply focus on \texttt{Accuracy} for evaluation.
For models that generate answers, we use their answer accuracy. For equation-generating models, we use a calculator to assess the accuracy of their equations as their answer accuracy.

\subsection{Overall Performance Comparison}

We compare our model, incorporated with dimension perception capability, with the selected LLMs on \uniteval.
Results are displayed in Table \ref{tab:Dimperc_result} and Table \ref{tab:dimperc_ablation}.

\subsubsection{RQ1}
From the data presented in Table \ref{tab:Dimperc_result}, we draw the following conclusions:
(1) \textit{Existing powerful Large Language Models (LLMs) possess an extensive understanding of basic unit knowledge, yet there remains a gap in dimension perception and scale perception.} 
Powerful LLMs like GPT-series perform well in the task of quantity extraction.
GPT-4 reaches an F1-score of 73.91\% in quantity extraction, capable of accurately identifying and extracting quantities in most cases.
The precision of the GPT-4 model in the quantity kind matching task reached 66.67\%. 
Even though this metric is not particularly high, it is close to the performance of our fine-tuned model, suggesting that the acquisition of such knowledge is relatively challenging.
Regarding dimension and scale perception, the GPT-series models exhibit relatively weaker performance.
GPT-4 can handle tasks that appear more frequently in natural language texts, such as unit conversion and comparable analysis, with F1-Score of 52.68\% and 55.18\%, respectively. 
However, its performance significantly diminishes when it comes to tasks involving knowledge-integrated reasoning, especially in unconventional tasks like dimension arithmetic, where its precision drops to 31.11\%.
(2) \textit{Smaller LLMs exhibit insufficient knowledge of quantities.}
Similar to powerful models, smaller models are more competent in tasks such as basic perception, unit conversion, and dimension perception, with LLaMa-2-13B achieving an F1-score of 36.62\% in dimension perception.
However, compared to our model, smaller models still have significant room for improvement in these tasks.
For more complex issues, smaller models are essentially incapable of providing solutions, indicating a lack of capacity in knowledge and reasoning tasks.
(3) \textit{Dimension perception is the most essential and challenging task in understanding units.} 
LLMs have relatively low accuracy in all three tasks in this aspect. 
Among them, the comparable task is actually the one most related to human daily activities, yet LLMs only achieve a precision up to 47.48\% in this task. 
This might be attributed to the fact that dimension-related tasks fall under the fundamental logic of quantitative understanding, necessitating a comprehensive grasp of dimensional knowledge and advanced logical reasoning abilities within the model.
(4) \textit{LLMs prefer to provide answers of high confidence while refraining from giving uncertain responses.}
We observe that the precision and F1-scores of LLMs are inconsistent across various tasks. 
Although these two metrics cannot be directly compared, the F1-scores of most models are significantly worse than the results indicated by the precision scores to a certain extent.
Through case studies, we find that even though our task setup involves multiple choice questions, LLMs still tend to refrain from providing responses to the questions they are unsure about or believe have no answer, which results in lower F1-scores. This situation is more pronounced in the GPT-series of models.

\subsubsection{RQ2}

Table \ref{tab:dimperc_ablation} shows the comparison of results between the baseline model and our model in dimension perception tasks.
It is evident that our model exhibits significant performance improvements across tasks in all categories. 
The precision for basic perception, dimension perception, and scale perception increase from 29.65\%, 20.38\%, 8.94\% to 71.69\%, 82.82\%, and 89.74\% respectively, while the F1-scores also show respective improvements.

The experimental results indicate that the base model, even after the instruction fine-tuning, remains deficient in basic knowledge, dimensional knowledge, and scale knowledge. 
By employing a dataset constructed based on \unitdb for fine-tuning, we effectively incorporate dimension knowledge and the corresponding capabilities into our model.

\begin{table}[t]
    \centering
    \caption{Accuracy(\%) of different models and settings on N-MWP and Q-MWP.}
    \label{tab:mwp results}

\resizebox{0.49\textwidth}{20mm} {
    \begin{tabular}{lcccc}
        \toprule

\multicolumn{1}{l}{Tasks} & \multicolumn{2}{c}{Numerical MWP} & \multicolumn{2}{c}{Quantitiative MWP} \\
\cmidrule(lr){2-3} \cmidrule(lr){4-5}
\multicolumn{1}{l}{Models} & \multicolumn{1}{c}{N-Math23k} & \multicolumn{1}{c}{N-Ape210k} & \multicolumn{1}{l}{Q-Math23k} & \multicolumn{1}{c}{Q-Ape210k}  \\ 
\midrule

\rowcolor[gray]{0.95} \multicolumn{5}{c}{\textit{Powerful LLMs}}\\

 \multicolumn{1}{l}{GPT4} & \multicolumn{1}{c}{78.22} & \multicolumn{1}{c}{65.33} & \multicolumn{1}{c}{\underline{57.33}} & \multicolumn{1}{c}{34.67} \\ 

 \multicolumn{1}{l}{\textit{\ \ + WolframAlpha}} & \multicolumn{1}{c}{\textbf{84.44}} & \multicolumn{1}{c}{\textbf{67.11}} & \multicolumn{1}{c}{{54.67}} & \multicolumn{1}{c}{\underline{43.55}} \\

 \multicolumn{1}{l}{GPT-3.5-turbo} & \multicolumn{1}{c}{49.33} & \multicolumn{1}{c}{{39.56}} & \multicolumn{1}{c}{{29.78}} & \multicolumn{1}{c}{14.22} \\ 

 \multicolumn{1}{l}{\textit{\ \ + WolframAlpha}} & \multicolumn{1}{c}{58.67} & \multicolumn{1}{c}{44.89} & \multicolumn{1}{c}{30.22} & \multicolumn{1}{c}{20.44} \\ 







\rowcolor[gray]{0.95} \multicolumn{5}{c}{\textit{Supervised Finetuned Models}}\\

 \multicolumn{1}{l}{BertGen} & \multicolumn{1}{c}{73.78} & \multicolumn{1}{c}{\underline{61.78}} & \multicolumn{1}{c}{{14.22}} & \multicolumn{1}{c}{{30.67}} \\ 
    
 
 \multicolumn{1}{l}{LLaMa} & \multicolumn{1}{c}{{78.22}} & \multicolumn{1}{c}{53.78} & \multicolumn{1}{c}{{36.44}} & \multicolumn{1}{c}{18.67} \\ 
 

    


\multicolumn{1}{l}{\textbf{DimPerc (Ours)}} & \multicolumn{1}{c}{\underline{80.89}} & \multicolumn{1}{c}{{60.00}} & \multicolumn{1}{c}{\textbf{82.67}} & \multicolumn{1}{c}{\textbf{50.67}} \\ 

        \bottomrule
\end{tabular}
}
\end{table}

\subsubsection{RQ3}
Table \ref{tab:mwp results} exhibits a comparative analysis between various baseline models and our model on the N-MWP and Q-MWP.
The results clearly show that LLMs such as GPT-3.5-turbo and GPT-4 perform significantly worse on Q-MWP compared to their performance on N-MWP.
This indicates that LLMs might not have fully grasped how to handle diverse units during pretraining, resulting in their less-than-ideal performance in quantitative reasoning tasks.
GPT-4 excels in handling N-MWP tasks, while GPT-3.5-Turbo continues to struggle with them, indicating that there is room for improvement in GPT-3.5-Turbo's logical reasoning and computational abilities.

The results also demonstrate that models fine-tuned on the current N-MWP dataset perform well in addressing N-MWP problems, yet it is still deficient in dealing with Q-MWP taks.
This reveals that the existing MWP datasets pay excessive attention to abstract values, whereas the focus on quantitative reasoning tasks encompassing diverse units is relatively insufficient.

On the Q-Ape210k dataset, tool-augmented GPT-4, the best-performing model without supervised finetuning, has an accuracy of 43.55\%.
The accuracy of the best-performing N-MWP trained model is 30.67\%. 
We elevate the problem-solving accuracy to 50.67\% by dimension perception enhancement.
Our approach not only significantly improves the performance on N-MWP tasks compared to LLMs under zero-shot scenarios, but also retains its effectiveness in solving the original N-MWP tasks.

\subsubsection{RQ4}
As can be seen from Table\ref{tab:Dimperc_result}, after the integration of \textit{WolframAlpha}, GPT-4 improved in tasks such as comparable analysis, dimension prediction, and scale perception, yet its performance in basic perception and dimension perception is declined.
The external knowledge and calculation capabilities of \textit{WolframAlpha} aid models in quantity reasoning to some extent, yet they fall short in comparison to our model.  
The disparity might be due to the fact that the current tool-model interfaces are not yet fully developed, and in comparison to \unitdb, their unit knowledge is less rich and diverse.
This highlights the superiority of our method in enhancing the quantitative reasoning capabilities of LLMs.

\begin{figure}[t]
    \centering
    \begin{minipage}{.225\textwidth}
        \centering
        \includegraphics[width=1\linewidth]{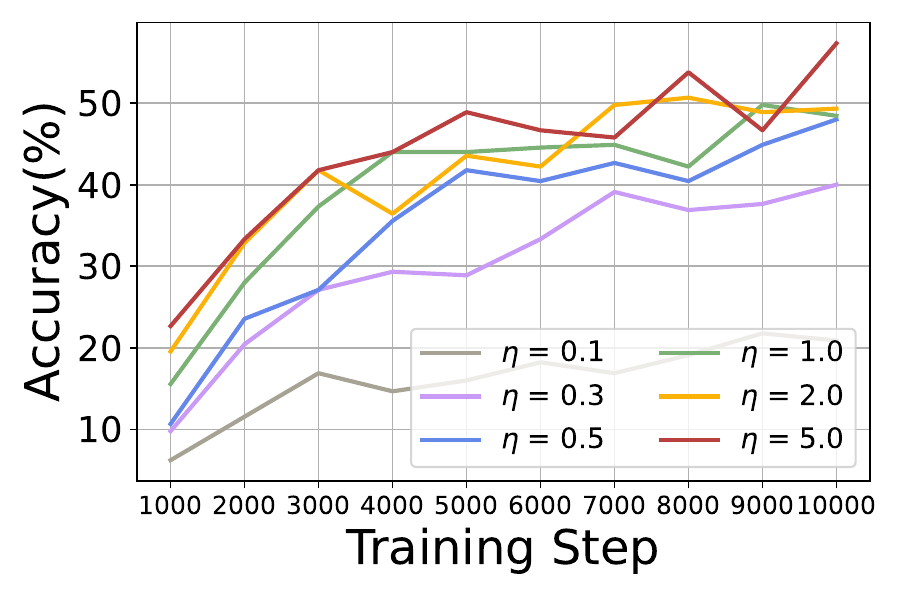}
        \caption{Accuracy of Dimperc model on Q-Ape210k with different data augmentation rate $\eta$.}
        \label{fig:quanres_dataauug}
    \end{minipage}
    \hfill
    \begin{minipage}{.225\textwidth}
        \centering
        \includegraphics[width=1\linewidth]{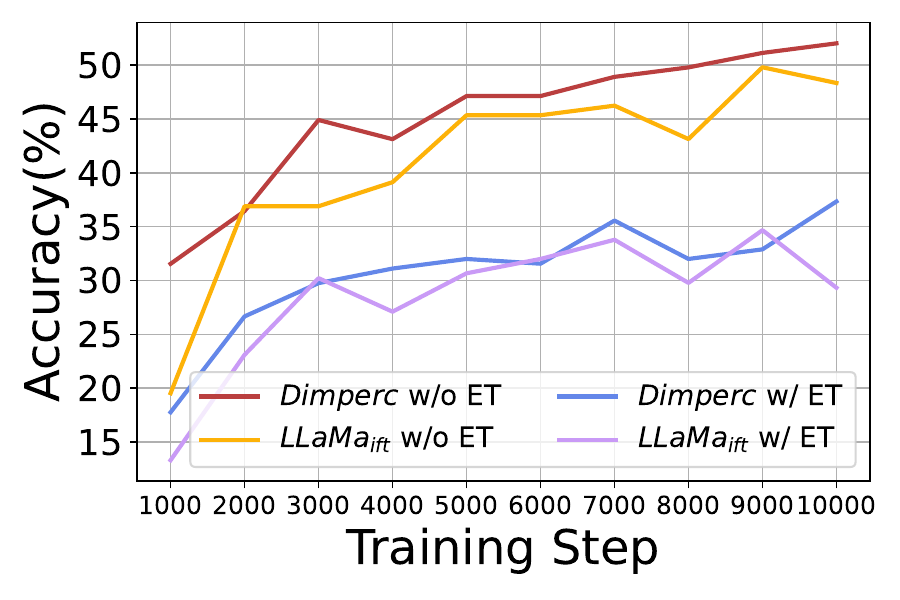}
        \caption{Accuracy of different base models on Q-Ape219k with different tokenization strategy.}
        \label{fig:quanrea_basemodel_et}
    \end{minipage}
\end{figure}

\subsection{Ablation Study}

We conduct experiments on Q-MWP using various base models and tokenization strategies. 
As depicted in Fig.~\ref{fig:quanrea_basemodel_et}, the \dimperc model, which is finetuned on \uniteval in Section~\ref{sec:dimperc}, exceeds the performance in quantitative reasoning tasks compared to its base model, particularly in the initial stages of training.
This indicates that finetuning on \uniteval task effectively incorporates dimension-related knowledge and skills into the language model. 
As the number of training steps increases, the performance of both base model and \dimperc gradually improves in quantity reasoning tasks.
This is because the finetuning process enables the models to learn quantity-related knowledge from the training data and become more adapted to mathematical reasoning tasks.

Furthermore, we observe that the performance of the equation tokenization is inferior to the regular tokenization strategy, contradicting prior research findings~\cite{ggb2020injecting}. 
This discrepancy might result from the improved numerical handling capabilities of the base model as its scale increases.
Additionally, an additional tokenization strategy introduces a gap between pretraining and finetuning, which further raises the fine-tuning cost for downstream tasks and impedes generalization.

\subsection{Influence of Important Hyper-parameters}

As indicated in Fig.~\ref{fig:quanres_dataauug}, we observe that increasing the proportion of data augmentation effectively enhances the model's performance on quantitative reasoning tasks. 
This is because a higher augmentation ratio adds more relevant knowledge, helping the model handle quantity units better.
Furthermore, we discover that for the Q-MWP task, configuring the data augmentation ratio to 0.5 or above yields beneficial outcomes. 
Considering both performance and training costs, setting $\eta$ to 0.5 is a suitable option.

\section{Related Work}

\subsection{Numerical Reasoning}
Numerical reasoning refers to the reasoning tasks that entail extensive processing of values.
DROP~\cite{dua-etal-2019-drop} proposes a benchmark that contains a large number of reading comprehension problems requiring numerical symbolic calculation to test the numerical understanding and reasoning ability of models. 
A series of problem-solving models for textual numerical reasoning problems are proposed and achieve good results on the DROP benchmark, such as NumNet~\cite{ran2019numnet}, GenBERT~\cite{ggb2020injecting}, NMNs~\cite{gupta2019neural}, etc.

Another common type of numerical reasoning is math word problems (MWPs)~\cite{xie2019goal}.
MWPs often take the form of real-world mathematical applications, involving arithmetic operations among values. They necessitate a higher level of mathematical reasoning, requiring complex and advanced knowledge and skills. 
These studies primarily focus on abstract values, overlooking values with units, leading to a one-sided understanding of values of existing language models.

\subsection{Quantitative Reasoning} 
Compared with numerical reasoning, quantitative reasoning is less studied.
At present, the research work in this area mainly includes measurement estimation and quantitative natural language reasoning~\cite{thawani-etal-2021-representing}.
Probing studies~\cite{zhang-etal-2020-language-embeddings,lin2020birds} on the DoQ dataset~\cite{elazar2019large} reveal the implicit knowledge of language models is not enough to realize quantity understanding, indicating a need for external knowledge.
UoM~\cite{park-etal-2022-language} pays attention to the quantity carrying units, and their probing experiments illustrate that language models lack the capability required for reasoning over measurements.

EQUATE~\cite{ravichander2019equate} applies quantity reasoning to the Recognizing Textual Entailment (RTE) task. 
WikiConvert~\cite{thawani-etal-2021-numeracy} constructs a cloze dataset to predict values and units in text.
These tasks primarily involve limited forms of values and units, and mostly require only single-step reasoning with minimal mathematical capacity demands.
In contrast, our quantitative reasoning tasks, featuring diverse unit representations and multi-step reasoning, are evaluated through math word problems that call for complicated mathematical skills.

\subsection{Tool Augmented Reasoning}
Several approaches employ external tools to enhance the reasoning capabilities of language models~\cite{gao2023pal, schick2023toolformer, paranjape2023art}.
ReAct~\cite{yao2022react} introduces a technique for collaborative large model inference and action, enabling models to ``reason" through a sequence of ``thought chains" and to ``act" by utilizing tools from a predefined toolkit, such as those capable of searching the internet.
WolframAlpha is a trending mathematical engine, capable of executing a variety of complex calculations in reasoning tasks, with a certain ability to handle unit conversion and quantitative processing.
Our method directly infuses the underlying logic of dimension perception and the ability for numerical reasoning into the model through finetuning.
\section{Conclusion and future works}

In this paper, we propose a framework to enhance the quantitative reasoning ability of LLMs based on dimension perception.
We construct a dimensional unit knowledge base (\unitdb), a comprehensive repository rich in units and their fundamental information. 
Based on this knowledge base, we develop a system that can handle units efficiently and conveniently.
We demonstrate the lack of dimensional knowledge in LLMs through evaluations on our proposed benchmark \uniteval.
Furthermore, we enhance the dimension perception skills of LLMs through finetuning.
Finally, we extend numerical reasoning tasks to quantitative reasoning tasks and conduct experiments. 
The results suggest that our method effectively enhances the model's capabilities in quantitative reasoning.

Despite the relative comprehensiveness and thoughtful design of \unitdb, there may arise a necessity to incorporate new units over time, to adapt to evolving demands and knowledge updates.
Finetuning for each database expansion is costly and inefficient.
Future work can focus on dimension perception methods that facilitate lightweight expansion.
Furthermore, in downstream experiments, due to the lack of datasets with diverse units, we only conduct mathematical reasoning tests on math word problems. 
However, the rich collection of units and dimensional attributes in our knowlegde base can also be applied to various fields such as biomedicine, finance, and engineering physics.
\section*{Acknowledgements}
This work is supported by Science and Technology Commission of Shanghai Municipality Grant (No. 22511105902) and Natural Science Foundation of China (No. 62102095).

\bibliographystyle{IEEEtran}
\bibliography{reference}

\end{document}